\def\year{2021}\relax
\documentclass[letterpaper]{article} 
\usepackage{aaai21}  
\usepackage{times}  
\usepackage{helvet} 
\usepackage{courier}  
\usepackage[hyphens]{url}  
\usepackage{graphicx} 
\usepackage{natbib}  
\usepackage{caption} 
\usepackage{subfigure}
\usepackage{multirow}
\usepackage{float}
\usepackage{cite} 
\usepackage{makecell}
\setcitestyle{authoryear,round}
\usepackage{amsmath}
\title{Is Each Layer Non-trivial in CNN? (Student Abstract) }
\author{
   Wei Wang,
        Yanjie Zhu,
        Zhuoxu Cui,  
        Dong Liang,
    \\
}
\affiliations{
     Center for Medical AI,Institute of Biomedical and Health Engineering, Shenzhen Institutes of Advanced Technology, Chinese Academy of Sciences\\

    Shen Zhen 518055, China \\
    wei.wang1@siat.ac.cn, dong.liang@siat.ac.cn, 86-755-86392243

}

\begin{document}

\maketitle

\begin{abstract}
Convolutional neural network (CNN) models have achieved great success in many fields. With the advent of ResNet, networks used in practice are getting deeper and wider. However, is each layer non-trivial in networks? To answer this question, we trained a network on the training set, then we replace the network convolution kernels with zeros and test the result models on the test set. We compared experimental results with baseline and showed that we can reach similar or even the same performances. Although convolution kernels are the cores of networks, we demonstrate that some of them are trivial and regular in ResNet.
\end{abstract}

\section{Introduction}
The structures of neural networks are getting more and more complex. There are two basis forms:  short-connection and no-connection. Short-connection:ResNet \cite{resnet}. No-connection: VGG \cite{2014arXiv1409.1556S}. Particularly, long-connection can be seen as a special no-connection in the local area. Long-connection: UNet \cite{unet}, SegNet \cite{7803544}. We say the layers are non-trivial if the performances change slightly after replacing the convolution kernel with zeros, vice versa. It is obvious that each layer in no-connection form is important. But, many layers in short-connection (ResNet) are trivial. 

The main contributions of this paper can be summarized as : We donate the non-trivial layers of ResNet are mainly concentrated on feature decomposition layers, which refers to the layers changing the number of channel dimension, when the model is over-parameterized.

\section{Analyze the Convolution Kernels of ResNet Replaced by $\mathbf{0}$}
ResNet residual unit can be formulated as:  
$$
\begin{aligned}
\mathbf{x}_{l+1} =& \sigma(\mathbf{x}_l+BN(\sigma(BN(\mathbf{x}_l*\mathbf{w}_l'))*\mathbf{w}_l''))\\
\mathbf{x}_{l+1} =& \sigma(BN(\mathbf{x}_l*\mathbf{w}^{1*1}_l)+BN(\sigma(BN(\mathbf{x}_l*\mathbf{w}_l'))*\mathbf{w}_l''))
\end{aligned}
$$
Replacing one of the convolution kernels in residual unit with $\mathbf{0}$ can be written as:  
$$
\begin{aligned}
\mathbf{x}_{l+1}' =& \sigma(\mathbf{x}_l+BN(\sigma(BN(\mathbf{x}_l*\mathbf{0}))*\mathbf{w}_l''))\\
=& \sigma(\mathbf{x}_l+BN(\sigma(\mathbf{\beta'})*\mathbf{w}_l''))\\
\mathbf{x}_{l+1}'' =&\sigma(\mathbf{x}_l+BN(\sigma(BN(\mathbf{x}_l*\mathbf{w}_l'))*\mathbf{0}))\\
=& \sigma(\mathbf{x}_l+\mathbf{\beta''})
\end{aligned}
$$
$$
\begin{aligned}
\mathbf{x}_{l+1}' =& \sigma(BN(\mathbf{x}_l*\mathbf{w}^{1*1}_l)+BN(\sigma(BN(\mathbf{x}_l*\mathbf{0}))*\mathbf{w}_l''))\\
=& \sigma(BN\mathbf{(x}_l*\mathbf{w}^{1*1}_l)+BN(\sigma(\mathbf{\beta'})*\mathbf{w}_l''))\\
\mathbf{x}_{l+1}^{\prime \prime} =& \sigma\left(B N\left(\mathbf{x}_{1} * \mathbf{w}_{l}^{1 * 1}\right)+B N\left(\sigma\left(B N\left(\mathbf{x}_{l} * \mathbf{w}_{l}^{\prime}\right)\right) * \mathbf{0}\right)\right) \\
=& \sigma\left(B N\left(\mathbf{x}_{l} * \mathbf{w}_{l}^{1 * 1}\right)+\beta^{\prime \prime}\right)\\
\mathbf{x}_{l+1}''' = &\sigma(BN(\mathbf{x}_l*\mathbf{0})+BN(\sigma(BN(\mathbf{x}_l*\mathbf{w}_l'))*\mathbf{w}_l''))\\
=&\sigma(\mathbf{\beta}''')+BN(\sigma(BN(\mathbf{x}_l*\mathbf{w}_l'))*\mathbf{w}_l''))
\end{aligned}
$$
$\mathbf{x}_{l},\mathbf{x}_{l+1}: $The input and output feature maps of the $lth$ residual unit.
$\mathbf{x}_{l+1}',\mathbf{x}_{l+1}'',\mathbf{x}_{l+1}''':$ The output feature maps of the $lth$ residual unit.
$\mathbf{w}_l',\mathbf{w}_l'':$ The first convolution kernel and the second convolution kernel of the $lth$ residual unit.
$*:$ Convolution operation.
$BN:$ Batch normalization operation.
$\mathbf{\beta}',\mathbf{\beta}'',\mathbf{\beta}''':$ The bias in BN layers.

\section{Experiment}
We chose ResNet34 and PSPNet-ResNet34 \cite{psp} to conduct a classification task and image segmentation task on Cifar-10 \cite{Krizhevsky} and T1 \cite{T1}, respectively. The baselines are 84\% and 87\%. We conducted three groups of experiments. Firstly, we replaced each layer's convolution kernel with 0, respectively (see Figure 1 in supplementary material). Secondly, except for the feature decomposition layers and adjacent layers, we replaced all the other convolution kernels with 0 in the one layer block which refers to a continuous layer with the same channel number (see Figure 2 in supplementary material). Thirdly, we replaced feature decomposition layers of short-connection with 0 (see Figure 3 in supplementary material).

\section{Results}

The classification results of Cifar-10 are shown in Figure 1 and Table 1,2. The segmentation results of T1 are shown in Figure 2 and Table 3,4.
\begin{figure}[!htbp]
\centering
\includegraphics[scale=0.3]{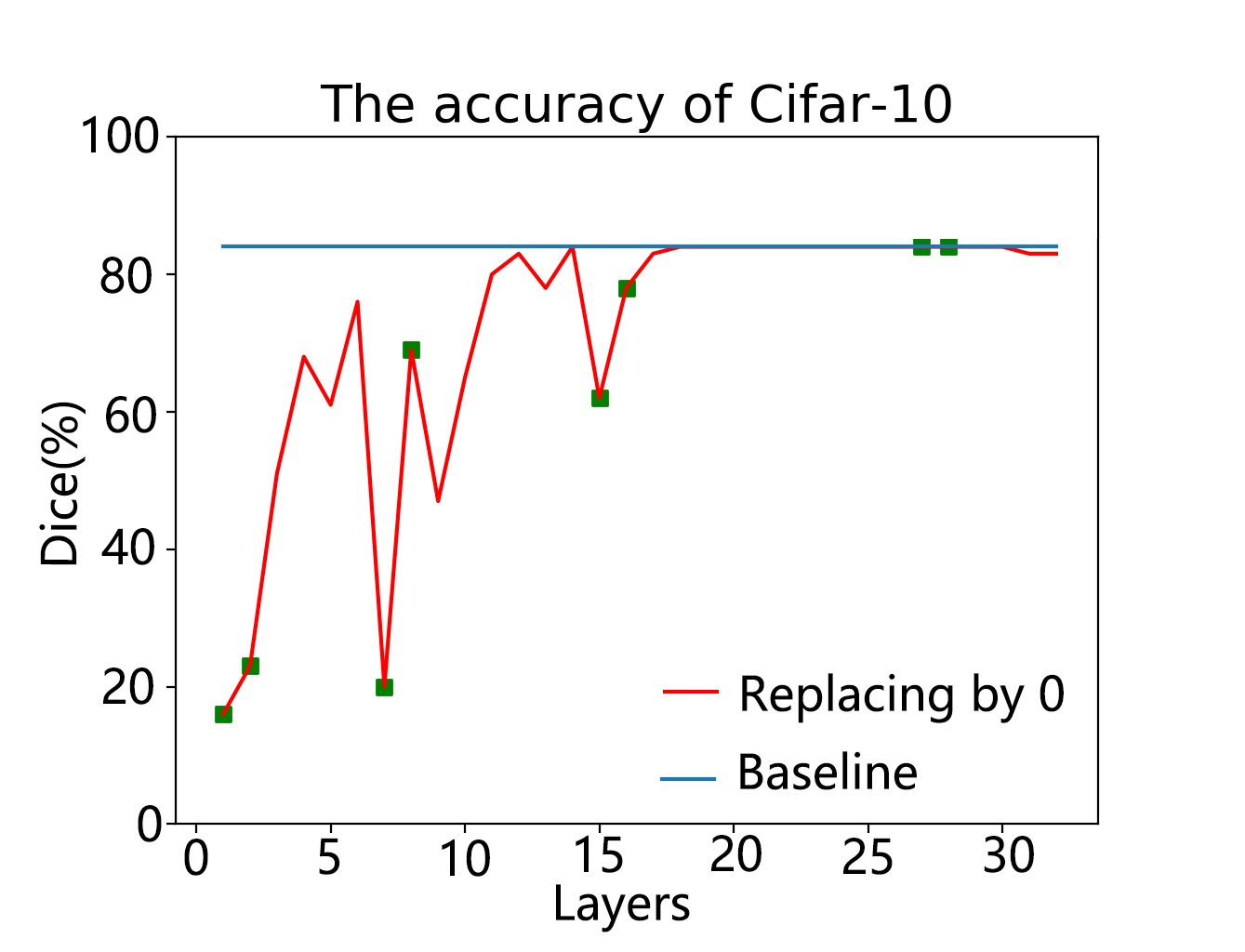}
\caption{The first experimental results for Cifar-10.}
\label{cifar}
\end{figure}  
\begin{table}[!htbp]
    
    \centering
    
    \begin{tabular}{|l|l|l|}
        Layer block  & ACC(\%) \\  
        Layer block 1  & 0.51 \\  
        Layer block 2  & 0.61 \\  
        Layer block 3  & 0.83 \\  
        Layer block 4  & 0.84 \\  
    \end{tabular}
\end{table}
\begin{table}[!htbp]
    \caption{The third experimental results for Cifar-10.}
    \centering
    \centering
    \begin{tabular}{|l|l|l|}
        Layer block  & ACC(\%) \\  
        Layer block 2  & 0.28 \\  
        Layer block 3  & 0.33 \\  
        Layer block 4  & 0.16 \\ 
    \end{tabular}
\caption{The second experimental results for  Cifar-10.}
\end{table}
\begin{figure}[!htbp]
\centering
\includegraphics[scale=0.3]{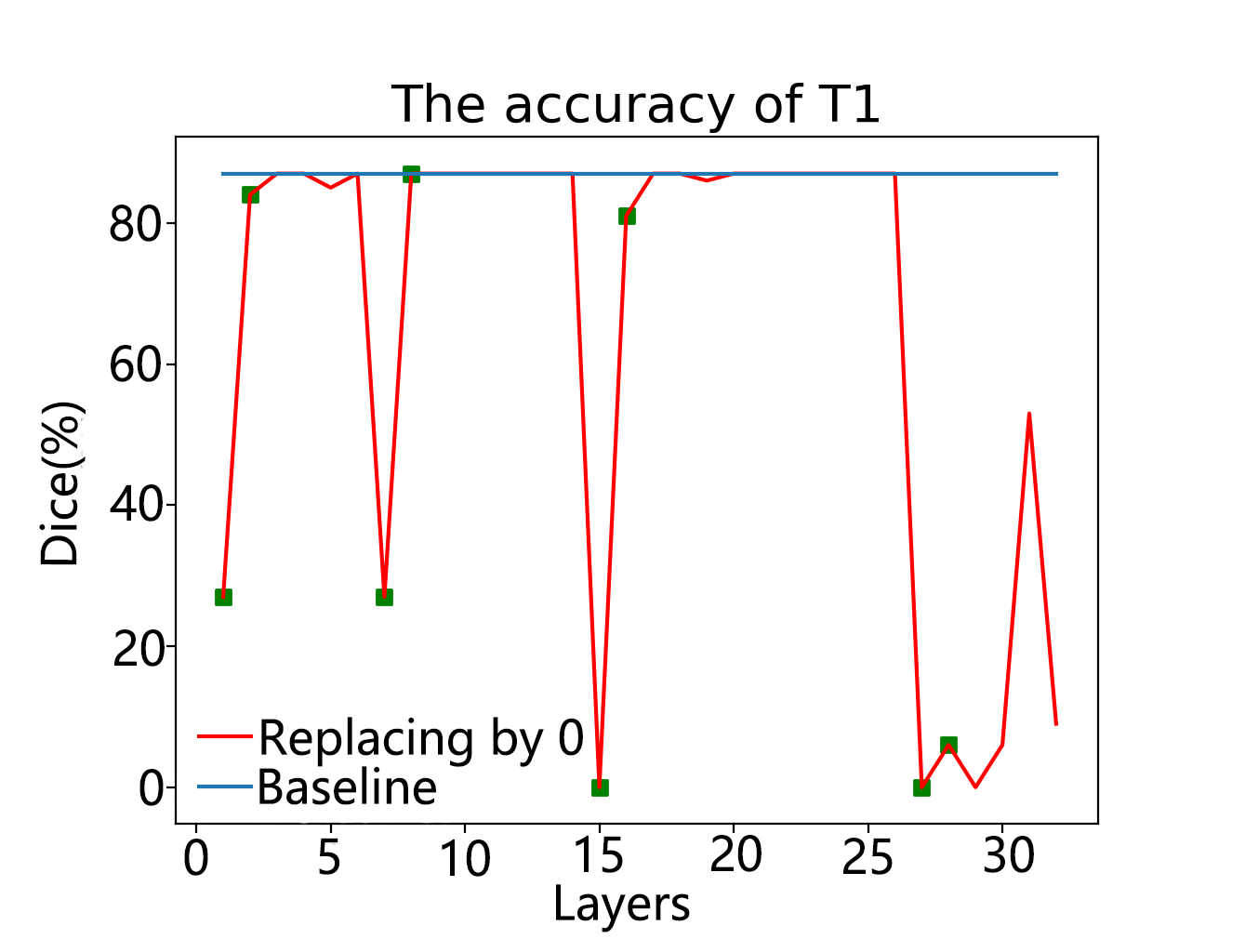}
\caption{The first experimental results for T1. }
\label{T1}
\end{figure}  
\begin{table}[!htbp]
    
    \centering
    \begin{tabular}{|l|l|l|}
        Layer block  & Dice \\  
        Layer block 1  & 0.82 \\  
        Layer block 2  & 0.86 \\  
        Layer block 3  & 0.82 \\  
        Layer block 4  & 0.00 \\  
    \end{tabular}
\caption{The second experimental results for T1.}
\end{table}
\begin{table}[!htbp]
    
    \centering
    \begin{tabular}{|l|l|l|}
        Layer block  & Dice \\  
        Layer block 2  & 0.00 \\  
        Layer block 3  & 0.00 \\  
        Layer block 4  & 0.00 \\ 
    \end{tabular}
   \caption{The third experimental results for T1.}
\end{table}

According to the structure of ResNet and the result of Figure 1 and Figure 2, it is obvious that the feature decomposition layers' convolution kernels are non-trivial, while the rests are trivial. Table 1 and Table 3 also confirm our conjecture. Table 2 and Table 4 also demonstrate the feature decomposition layers of short-connection are non-trivial.

\section{Discussion}
We argue that ResNet is a continuous process of feature decomposition and information storage. 
ResNet shows different changes in non-trivialness at the front and back of the network for different tasks. Generally, the classification task needs to learn enough information about the global abstract feature. Since enough information has learned in the front, the back layers are no longer non-trivial. Segmentation requires information for each pixel, so the back layers are non-trivial.

\section{Conclusion}

When there are redundant parameters in ResNet, not all layers of the network are non-trivial, or some layers may not be needed when the network parameters have learned enough information. The feature decomposition layers and identity mappings are important. Particularly, the feature decomposition layers are responsible for the feature decomposition, the identity mapping is responsible for the information storage and the residual layers are responsible for the adjustment of the feature to make it fit the final target. According to the above conclusions and experiments, when the model is over-parameterized, we can eliminate unnecessary layers in the ResNet and improve the training efficiency on the premise of ensuring performance.

\bibliography{ref}

\begin{thebibliography}{7}
\providecommand{\natexlab}[1]{#1}
\providecommand{\url}[1]{\texttt{#1}}
\providecommand{\urlprefix}{URL }
\expandafter\ifx\csname urlstyle\endcsname\relax
  \providecommand{\doi}[1]{doi:\discretionary{}{}{}#1}\else
  \providecommand{\doi}{doi:\discretionary{}{}{}\begingroup
  \urlstyle{rm}\Url}\fi

\bibitem[{{Badrinarayanan}, {Kendall}, and {Cipolla}(2017)}]{7803544}
{Badrinarayanan}, V.; {Kendall}, A.; and {Cipolla}, R. 2017.
\newblock SegNet: A Deep Convolutional Encoder-Decoder Architecture for Image
  Segmentation.
\newblock \emph{IEEE Transactions on Pattern Analysis and Machine Intelligence}
  39(12): 2481--2495.

\bibitem[{Fahmy et~al.(2019)Fahmy, El-Rewaidy, Nezafat, Nakamori, and
  Nezafat}]{T1}
Fahmy, A.~S.; El-Rewaidy, H.; Nezafat, M.; Nakamori, S.; and Nezafat, R. 2019.
\newblock Automated analysis of cardiovascular magnetic resonance myocardial
  native T1 mapping images using fully convolutional neural networks.
\newblock \emph{Journal of Cardiovascular Magnetic Resonance} 21(1).

\bibitem[{{He} et~al.(2015){He}, {Zhang}, {Ren}, and {Sun}}]{resnet}
{He}, K.; {Zhang}, X.; {Ren}, S.; and {Sun}, J. 2015.
\newblock {Deep Residual Learning for Image Recognition}.
\newblock \emph{arXiv e-prints} arXiv:1512.03385.

\bibitem[{Krizhevsky(2012)}]{Krizhevsky}
Krizhevsky, A. 2012.
\newblock Learning Multiple Layers of Features from Tiny Images.
\newblock \emph{University of Toronto} .

\bibitem[{{Ronneberger}, {Fischer}, and {Brox}(2015)}]{unet}
{Ronneberger}, O.; {Fischer}, P.; and {Brox}, T. 2015.
\newblock {U-Net: Convolutional Networks for Biomedical Image Segmentation}.
\newblock \emph{arXiv e-prints} arXiv:1505.04597.

\bibitem[{{Simonyan} and {Zisserman}(2014)}]{2014arXiv1409.1556S}
{Simonyan}, K.; and {Zisserman}, A. 2014.
\newblock {Very Deep Convolutional Networks for Large-Scale Image Recognition}.
\newblock \emph{arXiv e-prints} arXiv:1409.1556.

\bibitem[{{Zhao} et~al.(2016){Zhao}, {Shi}, {Qi}, {Wang}, and {Jia}}]{psp}
{Zhao}, H.; {Shi}, J.; {Qi}, X.; {Wang}, X.; and {Jia}, J. 2016.
\newblock {Pyramid Scene Parsing Network}.
\newblock \emph{arXiv e-prints} arXiv:1612.01105.

\end{thebibliography}
\bibliographystyle{plain}
\end{document}